# Multilingual Tokenization through the Lens of Indian Languages: Challenges and Insights

**Maharaj Brahma**[*1], **N J Karthika**[*2],
**Rajat Verma**[3], **Nagasai Saketh Naidu**[4†], **Rohit Saluja**[3],
**Maunendra Sankar Desarkar**[1], **Ganesh Ramakrishnan**[2]
[1]Department of CSE, Indian Institute of Technology Hyderabad,
[2]Department of CSE, Indian Institute of Technology Bombay,
[3]SCEE, Indian Institute of Technology Mandi, [4]Adobe Research
**Correspondence:** cs23resch01004@iith.ac.in, karthika@cse.iitb.ac.in

## Abstract

Tokenization plays a pivotal role in NLP and is fundamental to training language models. However, existing tokenizers are often skewed towards high-resource languages, limiting their effectiveness for linguistically diverse and morphologically rich languages such as those in the Indian subcontinent. In this work, we present a comprehensive empirical study of multilingual tokenization across 17 Indic languages spanning 11 scripts and two language families. We systematically evaluate the effects of (i) widely used subword algorithms: BPE (Sennrich et al., 2016) and Unigram LM (Kudo, 2018), (ii) script and orthography-aware normalization, (iii) vocabulary size, and (iv) multilingual vocabulary construction strategies. We use a combination of intrinsic and extrinsic evaluations to obtain the following observations: (i) script-specific normalization improves tokenization quality, (ii) Unigram LM better preserves morphological boundaries than BPE, (iii) cluster-based vocabulary construction shows improvement in downstream tasks compared to the joint method. Our findings highlight the importance of linguistically informed design choices in multilingual tokenization and offer practical guidance for building effective tokenizers for low-resource and morphologically complex languages.

## 1 Introduction

Tokenization is the process of segmenting raw text into smaller units/tokens (words, subwords, characters, etc.), which helps in efficient processing by the computational models, particularly in Large Language Models (LLMs). Tokenizer step forms a fundamental step in any Natural Language Processing (NLP) task, and hence the quality of the tokenization impacts the model accuracy and training speed. Poor tokenization can lead to an increase in sequence lengths, fragment meaningful units, and weaken model's ability to capture linguistic structure and semantics. Tokenization further influences how well a model understands the linguistic structure and semantics of the input and how well it handles the vocabulary coverage. Most widely used tokenizers are designed primarily for English, benefiting from abundant data and extensive research. As a result, they are optimized for English-specific linguistic properties and are often reused for Indic languages, despite their distinct morphological and script characteristics, leading to suboptimal performance for morphologically rich and script-diverse languages. This challenge becomes more prominent in multilingual settings. Unlike monolingual tokenizers, a multilingual tokenizer must allocate a single shared vocabulary across languages, which may vary across scripts, morphology, and resource availability.

Zouhar et al. (2023) and Ali et al. (2024) conduct an extensive study to understand the influence of tokenization with the help of various intrinsic and extrinsic evaluation metrics. While previous works like Rust et al. (2021); Limisiewicz et al. (2023) have addressed tokenizer evaluation in multilingual contexts, Indic languages have received relatively lesser attention. Indic languages in particular are interesting for tokenization studies due to their rich morphological structure, agglutinative and inflectional properties, and diverse scripts. Unlike English and other high-resource languages, Indic languages exhibit a complex word formation process, extensive inflectional paradigms, and orthographic variations. Furthermore, the disparity in resource availability across languages may lead to a disproportionate favoring of high-resource languages. To systematically study these challenges, we formulate the following research questions:

(**RQ1**) *What is the impact of script and orthographic normalization on tokenization quality for Indic languages?*

[*]Equal Contribution
[†]Work done during B.Tech. at IIT Bombay.

(**RQ2**) *How do different subword algorithms vary in their ability to preserve morphological alignment in Indic languages?*

(**RQ3**) *Which multilingual vocabulary creation strategies, such as joint or cluster-based methodologies, yield the most effective tokenization, and what is the optimal way to group languages?*

Given the rich morphological and lexical characteristics of Indic languages and the script diversity, it is crucial to study how well various tokenization algorithms are able to capture these characteristics effectively. To the best of our knowledge, this study is among the first large-scale evaluations specifically focusing on tokenization behavior across a typologically and script-wise diverse set of 17 Indic languages. In this work, we present various methods for tokenizer training and vocabulary building, with a focus on multilingual Indian languages, and perform evaluations to understand the tokenizer's ability to capture the aforementioned characteristics of these languages. We particularly focus on the most widely used tokenizers *viz.,* Byte Pair Encoding (BPE) and Unigram Language Model (ULM), with vocabulary sizes ranging from 32K to 256K. Our contributions are summarized as follows:

1. Investigating the performance and impact of multilingual tokenization for 17 Indic languages from 2 language families *viz.,* Indo-European and Dravidian, with 11 different writing systems.

2. We analyse the impact of language-specific script normalization on tokenization quality, vocabulary size, and multilingual vocabulary construction strategies, using both intrinsic metrics and downstream task performances. We further study two BPE variants: PickyBPE (Chizhov et al., 2024) and OBPE (Patil et al., 2022).

## 2 Related Works

**Subword tokenization.** Tokenization is a fundamental task in Natural Language Processing, with subword tokenization algorithms such as Byte Pair Encoding (BPE) (Sennrich et al., 2016) and Unigram Language Model (ULM) (Kudo, 2018) remaining the most widely adopted algorithms for NLP tasks. Additionally, recent work have introduced several new tokenization algorithm, which are mostly variants of BPE and ULM, such as Overlap BPE (Patil et al., 2022), PickyBPE (Chizhov et al., 2024), Scaffold BPE (Lian et al., 2025), BPE Knockout (Bauwens and Delobelle, 2024), and Conditional Unigram LM (Vico and Libovický, 2025). Several studies have investigated the impact of tokenization, focusing on various aspects including the effects of tokenization on model capacity, generalisation, and fairness across languages (Limisiewicz et al., 2023; Schmidt et al., 2024). Works examining these aspects from the context of Indian languages so far are limited.

**Multilingual tokenization.** Multilingual tokenizers face an inherent trade-off between enabling cross-lingual transfer and maintaining optimal monolingual performance. This challenge was highlighted by Rust et al. (2021), demonstrating that replacing a shared multilingual tokenizer with a language-specific one, can improve the downstream performance of a model. Subsequent research has focused on this challenge by improving on how shared vocabularies are allocated across languages. Chung et al. (2020) proposed a method to improve multilingual vocabulary allocation by clustering languages based on vocabulary overlap, while Limisiewicz et al. (2023) studied the impact of vocabulary overlap and allocation for 20 languages. More recently, Abagyan et al. (2025) found that training tokenizers with additional languages beyond the languages used in pretraining improves the adaptation capabilities to unseen languages. Building on these insights, our work investigates optimal tokenization strategies specific to morphologically rich and script-diverse Indian languages by conducting an empirical comparison of various strategies to determine their effectiveness specifically for Indian languages.

## 3 Subword Tokenization

Subword tokenization is a fundamental technique in modern NLP, particularly for LLMs. Recent work has highlighted the critical roles of tokenization in multilingual settings with implications for both model performance and token fairness (Petrov et al., 2023; Ali et al., 2024). This issue is especially pronounced for Indic languages, which covers a number of languages with diverse scripts, rich morphology, and limited representation in the pretraining corpus.

Most contemporary subword tokenization frameworks are based on Byte Pair Encoding (BPE) and Unigram Language Model (ULM) based approaches. Recently proposed tokenizers (Lian et al., 2025; Chizhov et al., 2024; Bauwens and Delobelle, 2024), whether specific to a language or multilingual, are essentially extensions or variants of these algorithms. Hence, we choose BPE and ULM for all experiments in this study as they provide coverage with respect to the dominant approaches and provide a comprehensive basis for evaluating the effects of multilingual vocabulary construction and varying vocabulary sizes.

### 3.1 Algorithms

In this section, we describe two widely used tokenization algorithms.

**Byte-pair encoding (BPE)** (Sennrich et al., 2016): BPE is a bottom-up subword-segmentation approach that iteratively merges the most frequent bigrams to build a vocabulary till the required vocabulary size is reached. It begins with Unicode characters as the initial vocabulary and greedily performs the merges.

**Unigram Language Model (ULM)** (Kudo, 2018): In contrast to BPE, ULM follows a top-down approach. It begins with the superset of the target vocabulary and iteratively prunes it down to the desired vocabulary size. Unlike BPE's purely frequency-based approach, ULM leverages a probabilistic language model, which enables sampling of multiple possible segmentations. ULM uses the Expectation-Maximization (EM) algorithm to estimate the probabilities of subword units.

### 3.2 Methods of Combining Vocabulary

In this section, we discuss different methods of combining vocabularies of languages to build a multilingual shared vocabulary. Consider a set of languages $\mathcal{L} = \{\mathcal{L}_1, \mathcal{L}_2, \cdots, \mathcal{L}_n\}$ and corresponding monolingual corpus $\mathcal{D}_i$ for language $\mathcal{L}_i$. To obtain a multilingual tokenizer, we investigate two approaches *viz.*, joint and cluster.

#### 3.2.1 Joint Method for Multilingual Tokenizer

In this method, a multilingual corpus is created by concatenating data from all languages. A shared vocabulary is then obtained by training the tokenizer on this data. This method is straightforward and widely used (Sennrich et al., 2016; Ramesh et al., 2022; Doddapaneni et al., 2023). However, it may disproportionally favor high-resource languages during training, leading to under-representation of tokens from low-resource languages. We construct the multilingual corpus by $\mathcal{M} = \text{CONCAT}(\mathcal{D}_1, \ldots, \mathcal{D}_n)$ and derive the vocabulary $\mathcal{V}_M = \mathcal{T}(\mathcal{M}, \mathcal{K})$, where $\mathcal{T}$ denotes a vocabularization algorithm (e.g., BPE, ULM) and $\mathcal{K}$ is desired vocabulary size.

**Algorithm 1** Cluster-based Multilingual Tokenizer

**Require:** **[Input]** Set of languages $L$, target vocab size $V_{total}$, number of clusters $K$, Monolingual corpora $D^l$ for $l \in L$
**Ensure:** **[Output]** Multilingual Tokenizer $\mathcal{V}_M$
  **for** language $l \in L$ **do**
    $V^l \leftarrow$Train Unigram LM tokenizer on $D^l$
  **end for**
  $V^L \leftarrow \bigcup_{l \in L} V^l$
  **for** each language $l \in L$ **do**
    Initialize $v_l \in \{0,1\}^{|V^L|}$
    **for** each subword index $i$ in $V^L$ **do**
      **if** $V^L[i] \in V^l$ **then**
        $v_l[i] \leftarrow 1$
      **end if**
    **end for**
  **end for**
  Train $K$-means clustering on $\{v_l\}$
  Obtain clusters $C \leftarrow \{C_1, C_2, \ldots, C_K\}$
  **for** each cluster $c_i \in C$ **do**
    Concatenate corpora in cluster $D^{C_i} \leftarrow \bigcup_{l \in C_i} D^l$
    Compute proportion $p_i \leftarrow \dfrac{\left|\bigcup_{l \in C_i} V^l\right|}{\sum_{j=1}^{K} \left|\bigcup_{l \in C_j} V^l\right|}$
    Set vocabulary size $|V^{C_i}| \leftarrow p_i \times V_{\text{total}}$
    Train tokenizer on $D^{C_i}$ with vocab size $|V^{C_i}|$
    Extract cluster vocabulary $V^{C_i}$
  **end for**
  $V^{multi} \leftarrow \bigcup_i V^{C_i}$
  $\mathcal{D}_{merge} \leftarrow \bigcup_{l \in L} D^l$
  **for** each token $s \in V^{multi}$ **do**
    Compute $prob(s) \leftarrow \dfrac{\text{count}(s \mid \mathcal{D}_{merge})}{\sum_{t \in V^{multi}} \text{count}(t \mid \mathcal{D}_{merge})}$
  **end for**
  $\mathcal{V}_M \leftarrow (V^{multi}, \{prob(s)\}_{s \in V^{multi}})$

#### 3.2.2 Cluster-based Method for Multilingual Tokenizer

In the cluster-based method, languages that are typologically or script-wise similar, are grouped into clusters. We train ULM tokenizers on monolingual corpora for each language. Separate tokenizers are trained for each cluster, and the resulting vocabulary is then merged to get a final multilingual vocabulary. This approach reduces over-segmentation in low-resource languages by preserving vocabulary in each cluster. We follow Chung et al. (2020) to form clusters. Initially, we train monolingual tokenizers for 17 languages using the ULM and take the union of all the vocabularies $U_v$. We then create a language-specific vector of size $|U_v|$ by marking entries with 1 if the token is present in its vocabulary; otherwise, we mark it as 0. We then train a $K$-means clustering algorithm using the vector as input. We then train individual tokenizers for each cluster and merge them to obtain the final vocabulary. The detailed algorithm is presented by Algorithm 1. We identify clusters in two ways: automatically using K-means, and manually based on language family groupings.

# 4 Experimental Setup

## 4.1 Languages and Datasets

We consider 17 Indic languages from two diverse language families with 11 different scripts. Languages considered are Hindi (hin), Assamese (asm), Bengali (ben), Konkani (gom), Gujarati (guj), Kannada (kan), Maithili (mai), Malayalam (mal), Marathi (mar), Nepali (nep), Oriya (ori), Punjabi (pan), Sanskrit (san), Sindhi (snd), Tamil (tam), Urdu (urd), and Telugu (tel).

**Dataset:** We utilize the Sangraha corpus (Khan et al., 2024), which offers high-quality verified data. We sample 10% of the verified data and retain only the languages with more than 10k rows, resulting in a selection of 17 Indic languages. Further, we exclude sentences containing more than ten words written in Roman script, as these are likely code-mixed or non-standard. To ensure a balanced multilingual training corpus, we follow the sampling strategy of (Conneau and Lample, 2019), with a temperature parameter $\alpha = 0.3$. Detailed data sampling process and statistics are presented in Appendix A. We apply normalization on the sampled corpus using the IndicNLP library[1] (Kunchukuttan, 2020), which standardizes Unicode characters and diacritics across Indic scripts.

## 4.2 Language Modeling

**Data Preprocessing:** We preprocess the Sangraha corpus (Verified and Synthetic), following standard BERT preprocessing settings[2]. We apply masked language modelling with a masking probability of 0.15, masking up to 20 tokens per sequence and using a duplication factor of 2 with a fixed random seed for reproducibility.

**Pre-training & Finetuning:** We follow a standard BERT-base architecture (Devlin et al., 2019) with 12 transformer layers, a hidden size of 768, 12 self-attention heads, and a feed-forward dimension of 3,072. The learning rate follows a polynomial decay schedule with linear warmup. Masked token selection follows the standard BERT strategy, where 80% of selected tokens were replaced with `[MASK]`, 10% were left unchanged, and 10% were replaced with random tokens. Training was conducted on A6000s (8 cores) for roughly 48 hours. We finetune the trained model on the training set of the IndicXtreme benchmark (Doddapaneni et al., 2023) for NLI and NER tasks using IndicXNLI and Naampadam datasets, respectively (More details are provided in Appendix C).

## 4.3 Evaluation

We assess tokenizer performance using both task-independent tokenizer properties (intrinsic evaluation) and downstream model performance (extrinsic evaluation). For intrinsic evaluation, we analyze the tokenizers on the Flores-200 dev set (Costa-Jussà et al., 2022) containing 997 sentences. For morphological alignment, we use the MorphScore dataset (Ali et al., 2024) for Gujarati and Tamil, and the dataset by Brahma et al. (2025) for Hindi and Marathi.

### 4.3.1 Intrinsic Evaluation Metrics

**Fertility (F)** (Ács, 2019; Ali et al., 2024): It is the measure of the average number of tokens a space-separated word is split into. If every word is mapped to exactly one token, the fertility score $= 1$. If a word is broken into two or more subwords/tokens, the fertility score $>1$.

**Character Per Token (CPT):** (Limisiewicz et al., 2023): It is the average number of characters in a token, and gives an idea of how granular the tokenization is. A low CPT indicates higher granularity, suggesting an aggressive split, while a higher CPT value indicates lower granularity.

**Morphological Alignment:** **MorphScore** (Arnett and Bergen, 2025) measures how well a tokenizer splits words in alignment with their actual morpheme boundaries. In this paper, we use a variant of morphscore, which assumes that a word may be tokenized into two or more segments, while also accounting for possible modifications at the segment boundaries arising from the underlying morphemes that compose the word. For each word, we calculate the percentage of correct segments in the tokenized output compared to the morpheme-based segments (ground truth).

$$\text{MorphScore} = \frac{\text{\# correct segments}}{\text{total \# segments in ground-truth}}$$

### 4.3.2 Extrinsic Evaluation

We pre-train an encoder-based transformer model from scratch using the Sangraha corpus (Khan et al., 2024), following the BERT architecture, using two different tokenizers - joint and cluster-based. We evaluate both zero-shot and fine-tuned variants on two downstream tasks: natural language inference (NLI) and named entity recognition (NER). For NLI, we use the standard IndicXNLI dataset (Aggarwal et al., 2022), and for NER, we use the Naamapadam dataset (Mhaske et al., 2023) from the IndicXtreme benchmark (Doddapaneni et al., 2023).

[1] https://github.com/anoopkunchukuttan/indic_nlp_library

[2] https://github.com/google-research/bert

| Algorithm | Vocab | | | | | | | |
|---|---|---|---|---|---|---|---|---|
| | 32k | | 64k | | 128k | | 256k | |
| | NN | N | NN | N | NN | N | NN | N |
| BPE | 2.190 | **2.137** | 1.926 | **1.872** | 1.717 | **1.664** | 1.568 | **1.517** |
| ULM | 2.148 | **2.093** | 1.895 | **1.838** | 1.695 | **1.663** | 1.575 | **1.530** |

Table 1: Average fertility scores reported across 17 languages (Joint). Here, NN and N represent non-normalized and normalized, respectively.

## 5 Experiments and Results

We train multilingual tokenizers for 17 languages belonging to Indic language families using standard and language script normalization, employing two widely used subword algorithms: Byte Pair Encoding (BPE) and the Unigram Language Model (ULM). To study the effect of multilingual vocabulary construction, we consider two training regimes: (i) a joint, where a single shared vocabulary is learned by concatenating data from all languages, and (ii) a cluster based setting, where languages are grouped either using data-driven clustering or by language family, separate vocabularies are learned per cluster and subsequently merged into a multilingual tokenizer. We train all tokenizers in our experiments using *SentencePiece* (Kudo, 2018) library. The details of the hyper-parameter settings used are presented in Table 10.

### 5.1 Impact of Normalization

To investigate the impact of Indic script-specific normalization for all languages considered, we trained tokenizers on both normalized and non-normalized corpora using the joint training approach. We apply a fifth case normalization rule (N J et al., 2025), to convert words with anusvāra (ं) into the corresponding nasal consonant for all languages, wherever applicable. This ensures a standardized training corpus with fewer character variations. This normalization aligns with the foundational phonetic principles outlined in the Aṣṭādhyāyī and also aligns with the concept of nasal assimilation[3] in linguistics.

For example, consider the word in Hindi, अन्दर (andara, inside). We can find the same word written/typed with slight orthographic variations, for example: Using anusvāra ंद (ṁda) instead of the use of the corresponding fifth case consonant न्द (nda). Such variation could lead to the same word-form (aṁdara) appearing as two different tokens (aṁdara and andara) while building a vocabulary, causing redundancy and inefficient usage of vocabulary capacity. The orthographic variations in text is a very common occurrence across Indian languages, driven by diverse writing, and non-standardised typing practices, often leading to multiple surface forms for the same word. The pictorial depiction of normalization in shown in Figure 3.

Table 1 presents the average fertility scores across 17 languages for the tokenizers trained. For both 128k and 256k vocabulary sizes, tokenizers trained on normalized corpora consistently yielded lower fertility scores compared to non-normalized data. For instance, with a vocabulary size of 128k, the BPE achieves a fertility score of 1.664, compared to 1.717 on non-normalized data. For a tokenizer trained using the ULM normalized corpus, the model achieves a better fertility score of 1.663 as compared to 1.695 for the non-normalized corpus. Similar observations are made for tokenizers trained on a 256k vocabulary. The detailed fertility scores for non-normalized and normalized training corpora are presented in Table 16, Appendix E.1.1.

To understand the extent of language-script normalization, we estimated the proportion of words in a widely used dataset that undergo the normalization explained above. In the evaluation set of Flores-200 dev (Costa-Jussà et al., 2022) for languages we have considered, there are a total of 312,500 words, of which **4.36%** (13,621) words undergo changes with fifth-case normalization. Considering the percentage of unique words that undergo normalization changes, upto **6%** of total unique words undergo normalization. The language-wise statistics of the normalization effect are shown in Table 15.

Table 2 reports fertility scores for non-normalized and normalized tokenizers across all vocabulary sizes, along with absolute and relative reductions. We compute per-language fertility deltas and conduct a Wilcoxon signed-rank test across 17 languages; $p$-values are reported in Table 3. All reductions are statistically significant ($p < 0.05$). These findings show that although the fertility reduction is modest, the effect is consistent across languages and vocabulary sizes, suggesting that orthographic normalization systematically improves tokenizer efficiency.

[3]Nasal assimilation is a phonological process in which a nasal sound changes to match the place of articulation of the consonant that follows it.

| | BPE | | | | ULM | | | |
|---|---|---|---|---|---|---|---|---|
| **Vocab** | **NN** | **N** | **Δ** | **%** | **NN** | **N** | **Δ** | **%** |
| 32k | 2.190 | 2.137 | 0.053 | 2.42 | 2.148 | 2.093 | 0.055 | 2.56 |
| 64k | 1.926 | 1.872 | 0.054 | 2.80 | 1.895 | 1.838 | 0.057 | 3.01 |
| 128k | 1.717 | 1.664 | 0.053 | 3.09 | 1.695 | 1.663 | 0.032 | 1.89 |
| 256k | 1.568 | 1.517 | 0.051 | 3.25 | 1.575 | 1.530 | 0.045 | 2.86 |

Table 2: Fertility scores of non-normalized (NN) vs. normalized (N) tokenizers, with absolute delta (Δ) and relative reduction (%) per vocabulary size.

| **Algorithm** | **32k** | **64k** | **128k** | **256k** |
|---|---|---|---|---|
| BPE | 0.0048 | 0.0033 | 0.0003 | 0.0008 |
| ULM | 0.0106 | 0.0026 | 0.0091 | 0.0198 |

Table 3: Wilcoxon signed-rank test $p$-values across 17 languages. All values significant at $p < 0.05$ (Wilcoxon signed-rank, $n$=17).

## 5.2 Vocabulary size

Figure 1 shows the percentage of vocabulary overlap for BPE and ULM tokenizers, for various scripts under consideration, across different vocabulary sizes. Interestingly, the percentage overlap increases consistently from 32k to 128k, but drops for the 256k vocabulary. This trend holds across all scripts considered. Similar trend is also observed for different tokenizer pairs (Refer Table 18).

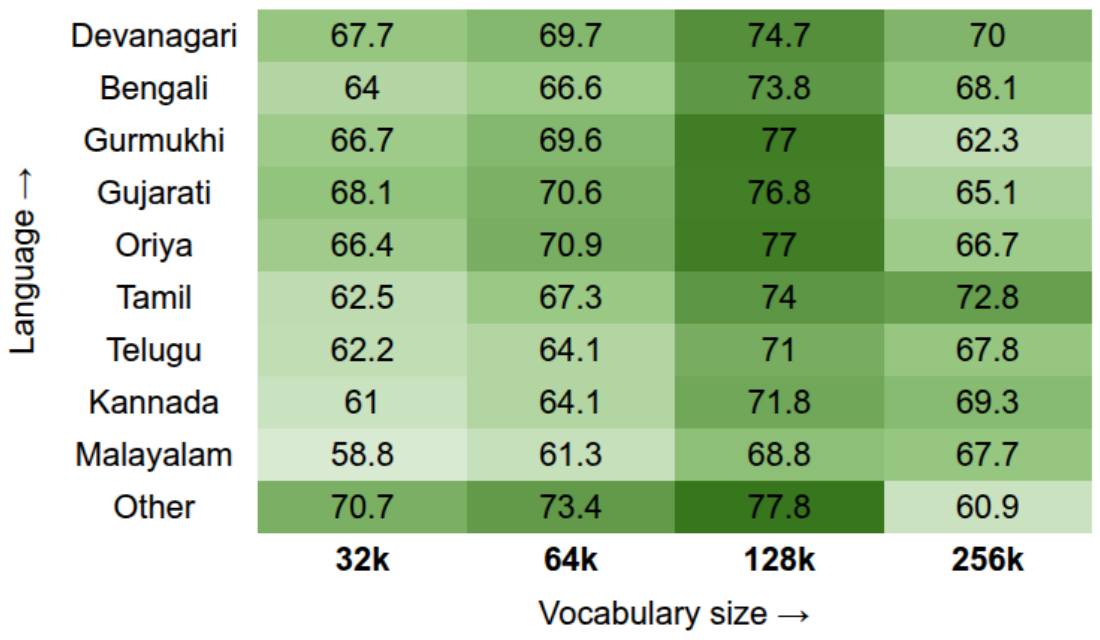

Figure 1: Overlap (%) between BPE and ULM vocabularies across scripts and vocabulary sizes.

We hypothesize that this change in trend occurs because 128k vocabulary captures most of the frequent tokens from the languages under consideration. After this point, the inclusion of additional tokens becomes increasingly arbitrary, which leads to a decrease in the overlap across independently trained tokenizers. Ideally, a robust multilingual tokenizer should produce a relatively consistent number of tokens for parallel sentences across different languages, although minor variations are expected due to linguistic differences. Considering this, we use the variance of the total token count of all languages as another evaluation metric to measure how consistently the tokenizer segments parallel content representing the same concept. Table 4 shows variance using 2 metrics *viz.,* Gini coefficient and normalized variance.

| **Algorithm** | **Vocab** | **Gini Coeff.** | **Normalized Variance** |
|---|---|---|---|
| BPE | 32k | 0.039 | 0.071 |
| | 64k | 0.034 | 0.063 |
| | 128k | 0.034 | 0.063 |
| | 256k | 0.042 | 0.073 |
| ULM | 32k | 0.038 | 0.069 |
| | 64k | 0.033 | 0.062 |
| | 128k | 0.036 | 0.065 |
| | 256k | 0.045 | 0.078 |

Table 4: Token count variance of BPE and ULM algorithm across various vocabulary sizes.

***Findings***: *Vocabulary overlap increases with vocabulary size upto 128k and then decrease at 256k. This suggests that tokens added become more random beyond 128k setting. This observation is further supported, to some extent, by the variance-based metrics.*

## 5.3 Morphological alignment

A primary challenge in evaluating morphological alignment for the diverse set of Indic languages is the scarcity of high-quality, standardized datasets with morpheme-level segmentations. To conduct a meaningful analysis under these constraints, we adopt two sets of best available resources. For Hindi and Marathi, we leverage the rich, multi-morpheme dataset provided by Brahma et al. (2025), which reflects the complex morphology of Indian languages, by segmenting the words into prefix(es), lemma, and suffix(es). We use a MorphScore variant as described in Section 4.3 against this multi-segmented ground truth. To expand our analysis, we use the Gujarati and Tamil language dataset presented by Ali et al. (2024), based on binary segmentation of the words. While such a binary split may oversimplify complex word structures in morphologically rich languages, yet it provides a consistent ground truth valuable for assessing the relative performance between BPE and ULM.

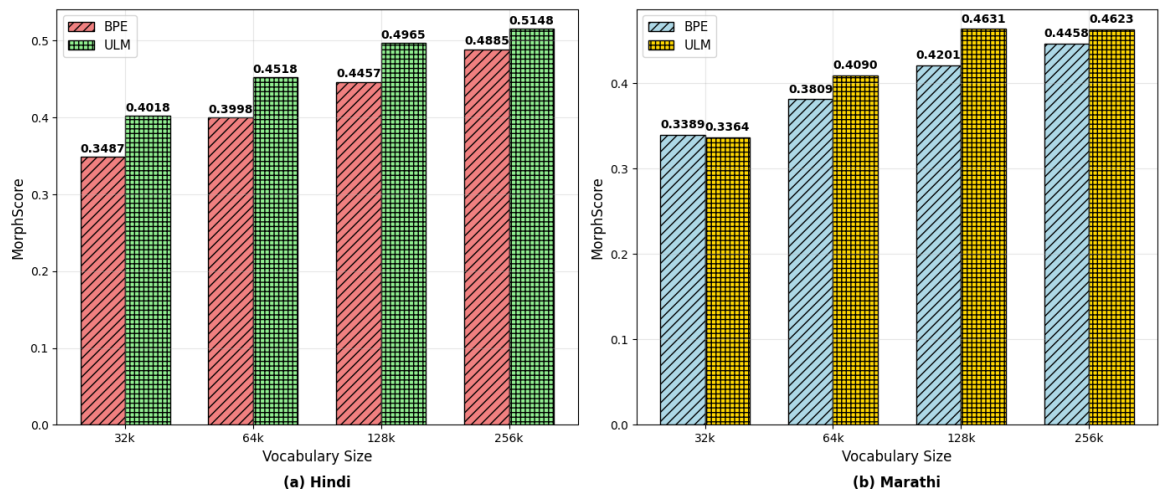

(a) MorphScore for Hindi and Marathi.

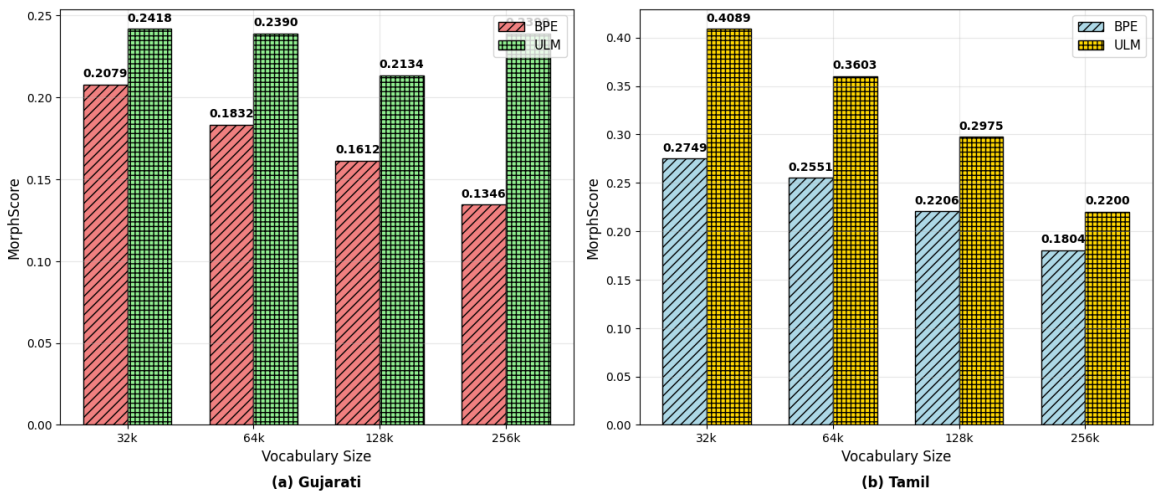

(b) MorphScore for Gujarati and Tamil.

Figure 2: Comparison of morphological alignment metrics across Indic languages. (a) MorphScore for Hindi and Marathi with varying vocabulary sizes. (b) MorphScore for Gujarati and Tamil.

Figure 2a shows the MorphScore for Hindi and Marathi data, which improves with the increase in vocabulary size. We also see that ULM consistently outperforms BPE in all four languages, which is in line with the findings by Bostrom and Durrett (2020). Figure 2b shows the MorphScore for Gujarati and Tamil data. In contrast to the trend shown with the multi-segmented data, here the morphscore decreases with the increase in vocabulary size. We hypothesize that this counter-intuitive result may be a consequence of the simplistic binary segmentation. Morphologically rich and agglutinative language like Tamil may have words made of multiple subwords fused together; for example, a word may be correctly split as m1+m2+m3. Since the dataset forces a single binary split, the ground truth may include m1+m2m3, ignoring the other valid morpheme boundary. With an increase in vocabulary size, the tokenizer's output may cater to more well-defined subwords, which doesn't match this binary split, causing the score to be lowered.

***Findings***: *Results suggest that ULM adheres more to the morphological segmentation compared to BPE, which is in line with the findings by Bostrom and Durrett (2020). We observe a reverse trend for Tamil and Gujarati with MorphScore decreasing as the vocabulary size increases. This trend is more prominent for ULM than for BPE. It may be due to the non-representative nature of the datasets for morphologically rich languages, where a word may be made of a number of subwords, whereas the words are split into only 2 segments in the dataset.*

### 5.4 Joint vs. Cluster

In the native script setting, the cluster variants (See Appendix D for details of clusters formed) achieve comparable or slightly lower fertility than the Joint method. The average fertility decreases marginally from 1.663 (joint) to 1.650−1.658 across the Cluster methods, indicating slight improvements. Notably, $\text{Cluster}_6$ attains the lowest score among the cluster variants. Additionally, we observe that as the cluster size increases, the fertility increases.

| Lang. | Fertility | | | | | |
|---|---|---|---|---|---|---|
| | Joint | $\text{Cluster}_5$ | $\text{Cluster}_6$ | $\text{Cluster}_7$ | $\text{Cluster}_8$ | $\text{Cluster}_{lf}$ |
| asm | 1.742 | 1.751 | 1.764 | 1.757 | 1.751 | **1.710** |
| ben | 1.489 | 1.496 | 1.506 | 1.495 | 1.496 | **1.466** |
| gom | 1.813 | 1.821 | 1.830 | 1.843 | 1.843 | **1.789** |
| guj | 1.570 | 1.574 | 1.562 | 1.578 | 1.610 | **1.540** |
| hin | 1.300 | 1.317 | 1.310 | 1.320 | 1.319 | **1.283** |
| kan | 2.225 | 1.990 | **1.968** | 1.989 | 1.982 | 2.126 |
| mai | 1.400 | 1.386 | 1.388 | 1.392 | 1.393 | **1.385** |
| mal | 2.240 | **2.165** | **2.165** | **2.165** | 2.166 | 2.356 |
| mar | 1.574 | 1.598 | 1.600 | 1.611 | 1.610 | **1.549** |
| nep | 1.560 | 1.573 | 1.574 | 1.585 | 1.585 | **1.528** |
| ori | 1.675 | 1.659 | 1.643 | 1.662 | 1.663 | **1.615** |
| pan | 1.424 | 1.419 | 1.409 | 1.420 | 1.421 | **1.392** |
| san | 1.975 | 2.000 | 2.046 | 2.061 | 2.064 | **1.954** |
| snd | 1.385 | 1.383 | 1.381 | 1.377 | 1.379 | **1.366** |
| tam | 1.837 | 1.846 | **1.831** | 1.857 | 1.847 | 1.950 |
| tel | 1.869 | 1.874 | **1.851** | 1.876 | 1.870 | 1.990 |
| urd | 1.197 | 1.199 | 1.194 | 1.192 | 1.193 | **1.182** |
| **Avg.** | **1.663** | **1.650** | **1.648** | **1.658** | **1.658** | **1.658** |

Table 5: Fertility scores for joint and cluster multilingual vocabulary construction with 128k vocabulary size. Highlighted and underlined values indicate the best scores.

To further investigate whether grouping related languages improves fertility, we manually constructed clusters based on language families ($\text{Cluster}_{lf}$). We notice that such a linguistically motivated clustering outperforms the automatically formed clusters for a majority of languages. Quantitatively, 13 out of 17 languages achieve lower fer-

| Language | IndicXNLI | | Naamapadam | |
|---|---|---|---|---|
| | Joint | $\text{Cluster}_{lf}$ | Joint | $\text{Cluster}_{lf}$ |
| asm | **26.18** | 21.30 | **3.01** | 2.03 |
| ben | **26.44** | 22.07 | 5.96 | **6.53** |
| guj | **26.85** | 22.78 | **7.18** | 4.79 |
| hin | **26.90** | 24.25 | **5.17** | 4.38 |
| kan | **26.87** | 24.09 | 5.41 | **7.45** |
| mal | **26.97** | 25.28 | **7.10** | 6.30 |
| mar | **26.69** | 24.14 | 6.81 | **6.95** |
| ori | **24.97** | 21.93 | 2.50 | **3.62** |
| pan | **26.78** | 21.59 | **5.97** | 4.97 |
| tam | **26.70** | 22.36 | 5.72 | **6.59** |
| tel | **26.90** | 22.77 | **6.32** | 6.15 |
| **Avg.** | **26.66** | 22.96 | **5.65** | 5.49 |

(a) Zero-Shot F1 scores (%) across 11 Indic languages on the Indic XNLI (NLI) and Naamapadam (NER) datasets.

| Language | IndicXNLI | | Naamapadam | |
|---|---|---|---|---|
| | Joint | $\text{Cluster}_{lf}$ | Joint | $\text{Cluster}_{lf}$ |
| asm | 63.98 | **64.93** | 48.00 | **56.25** |
| ben | **66.14** | 65.91 | **82.06** | 81.47 |
| guj | 64.12 | **65.12** | **82.77** | 81.24 |
| hin | 67.05 | **67.92** | 82.60 | **83.56** |
| kan | 64.08 | **65.80** | 84.51 | **84.65** |
| mal | 62.29 | **64.69** | **84.20** | 83.58 |
| mar | 63.52 | **65.00** | 82.68 | **83.73** |
| ori | 64.05 | **65.36** | 37.64 | **37.79** |
| pan | 65.44 | **67.15** | 74.83 | **75.92** |
| tam | 63.61 | **65.22** | 73.27 | **76.29** |
| tel | 63.58 | **64.99** | **84.44** | 84.40 |
| **Avg.** | 64.42 | **65.65** | 74.27 | **75.35** |

(b) Fine-tuned F1 scores (%) across 11 Indic languages on the Indic XNLI (NLI) and Naamapadam (NER) datasets.

Table 6: Extrinsic evaluation of joint and cluster-based multilingual tokenizers on downstream NLI and NER tasks.

tility scores than the joint method, indicating that language family-based grouping facilitates more effective vocabulary learning. However, we observe that three out of four languages belonging to the Dravidian language family have higher fertility than the joint. Suggesting that there is a trade-off among languages in vocabulary allocation.

***Findings***: *Table 5 suggests that cluster-based tokenizers better preserve language-specific subword units by mitigating the dominance of high-resource languages during vocabulary construction. Cluster-based vocabulary construction of multilingual tokenization has its own merit, with the reduction of the fertility and fairer splits as compared to the joint method. However, it seems to be sensitive to the formation of clusters, and careful consideration of languages per cluster is warranted.*

### 5.5 Extrinsic Evaluation

Table 6a & 6b summarizes the performance on the NLI and NER tasks under both zero-shot and fine-tuned settings. In the zero-shot setting, the joint vocabulary construction consistently outperforms the language family-based cluster vocabulary construction, yielding an average F1 score of +3.70 on IndicXNLI and +0.16 on Naamapadam. In contrast, under fine-tuning, the cluster-based vocabulary constructions show consistent gains across the majority of languages for both tasks. Specifically, language family (lf) based clustering achieves average improvements of +1.23 F1 on NLI and +1.08 on NER tasks compared to the joint vocabulary setting.

Overall, these results suggest that constructing vocabularies using language family-based clustering yields superior downstream performance in the fine-tuned setting, suggesting that learning a shared vocabulary via naive concatenation of multilingual data may be suboptimal, and that incorporating grouping of languages based on clusters during vocabulary construction can yield more effective token representations.

## 6 Analyses

### 6.1 Effect of Cluster-based method in Morphological alignment

Table 7 compares the Joint and Cluster variants for morphological alignment. We observe that clustering languages by language family ($\text{Cluster}_{lf}$) yields slightly higher or comparable scores than the Joint method for Tamil, Hindi, and Marathi. Although the gains are modest, these results suggest that the cluster-based multilingual vocabulary construction respects morphology more than the joint method.

| Lang. | MorphScore (↑) Joint | $\text{Cluster}_6$ | $\text{Cluster}_{lf}$ |
|---|---|---|---|
| guj | **0.213** | **0.213** | 0.208 |
| tam | 0.298 | 0.309 | **0.355** |
| hin | 0.497 | 0.480 | **0.504** |
| mar | 0.463 | 0.442 | **0.465** |

Table 7: Comparison of Joint vs. Cluster variants on morphological alignment using MorphScore.

### 6.2 Variant of BPE

**BPE vs. OBPE (Patil et al., 2022)**: Table 8 compares the BPE and Overlap-based BPE (OBPE) across six Indian languages with a vocabulary size of 64k[4] in the Devanagari script. These results highlight the limitations of the BPE algorithm in multilingual settings, while OBPE overcomes these limitations by generating a vocabulary that maximizes transfer from HRLs to related LRLs through token overlap between LRLs and related HRLs, while maintaining a compact corpus representation. **BPE vs. PickyBPE (Chizhov et al., 2024)**: Across all 17 languages, PickyBPE[5] (PBPE) consistently reduces fertility compared to BPE. On average, fertility decreases from 1.664 to 1.609 (small difference of 0.055). Suggesting that PBPE, though designed to reduce vocabulary bloating, shows a reduction in fertility (Refer to Appendix, Table 19).

[4]Due to computational cost, we perform experiments on six languages with a 64k vocabulary size. OBPE takes substantially higher training time than BPE.

| Algorithm | Languages | | | | | | Avg. |
|---|---|---|---|---|---|---|---|
| | gom | hin | mai | mar | nep | san | |
| BPE | **1.739** | 1.261 | 1.334 | **1.520** | **1.493** | 1.876 | 1.537 |
| OBPE | 1.747 | **1.255** | **1.326** | 1.521 | 1.494 | **1.875** | **1.536** |

Table 8: Fertility scores across six languages for OBPE and BPE with vocabulary size of 64k.

## 7 Conclusion

In this work, we perform a comprehensive study on the multilingual tokenization for Indic languages. We systematically examine the impact of script normalization across diverse scripts, existing subword algorithms (BPE and ULM), and vocabulary construction strategies. Our analysis reveals that cluster-based vocabulary construction consistently yields more efficient tokenizations and leads to improved performance in fine-tuned downstream tasks compared to joint vocabulary construction. While joint vocabularies show better performance in zero-shot settings, our results highlight the importance of incorporating linguistic structure when designing multilingual tokenizers. Furthermore, we show that script and orthographic normalization substantially influence tokenization quality, underscoring the need for careful preprocessing. Overall, our findings offer practical insights for designing effective multilingual tokenizers for underrepresented languages. While our focus is on Indic languages, the methodologies and insights are broadly applicable to other low-resource and morphologically complex language settings and contribute towards region-specific LLM programs like Ng et al. (2025); Gala et al. (2024).

## Limitations

Our study is limited to languages from distinct typological families and writing scripts, primarily spoken in India. Larger set of languages should be evaluated to understand the tokenization effect belonging to diverse language families. In this work, we evaluate the tokenization strategies on two downstream tasks: natural language inference and named entity recognition, and evaluation on more diverse tasks is left for future work. While our evaluation focuses on the performance of multilingual tokenizers using intrinsic metrics and extrinsic metrics, the influences of cross-lingual transfers among languages remain unexplored (Hämmerl et al., 2025).

While recent works (Arnett and Bergen, 2025; Vemula et al., 2025) suggest that morphological alignment and downstream performance may not be highly correlated, we nevertheless treat morphological alignment as a linguistically motivated intrinsic evaluation metric, particularly relevant to morphologically rich and agglutinative Indian languages. Lack of morphological alignment datasets for Indic languages restricted our analysis to four languages, highlighting the need for benchmark resources to better study the impact of morpheme-level tokenization over the alternatives.

More comprehensive extrinsic evaluation of training multilingual models with different cluster sizes, vocabulary sizes, varying parameter budgets, and diverse pretraining objectives would provide deeper insights into how tokenization design choices influence the representational quality of vocabulary and downstream task performance.

## Ethical Considerations

We do not identify any ethical risks based on this work. We used generative AI (ChatGPT) for language editing purposes only, such as paraphrasing, spell-checking, and refining and polishing the authors' original content.

## Acknowledgements

We express our deep appreciation to the reviewers for their invaluable and insightful suggestions, which greatly improved the quality of the paper. We also thank BharatGen (https://bharatgen.com/), IIT Bombay, IIT Hyderabad, and IIT Mandi for providing computational resources and support for this work. Furthermore, Karthika acknowledges TCS Research Foundation for supporting her research through a PhD fellowship.

[5] We do not compare OBPE vs. PickyBPE as they are not comparable directly, OBPE requires data to be in a similar script.

## A Tokenizer Training Corpus

We initially sampled data from the IndicCorpv2 corpus. However, upon manual inspection, we observed several language subsets included sentences in other languages and in different scripts. For instance, the Bodo corpus occasionally contained Hindi sentences. This is likely due to the quality of the filtering technique used to create the dataset. We thereafter decide to work on the Sangraha corpus (Ramesh et al., 2022).

Table 9 shows the detailed statistics of the tokenizer training corpus, totaling 39GB of data with 7.46M rows.

The equation for data sampling is presented below:

$$q_i = \frac{f_i^\alpha}{\sum_{j=1}^{N} f_j^\alpha} \qquad f_i = \frac{n_i}{\sum_{k=1}^{N} n_k}$$

Here, $n_i$ denotes the number of sentences in language $i$, $\alpha$ is the temperature parameter, and $q_i$ is the probability of sampling a sentence from that language.

## B Tokenizer

In our experiments, we use *SentencePiece* library (Kudo and Richardson, 2018). The settings we used are listed in Table 10. The settings that are not presented in the Table 10 are set to their default values.

For PickyBPE, we use the implementation of `https://github.com/bauwenst/PickyBPE`, and for OBPE, we use `https://github.com/Vaidehi99/OBPE`. We set the threshold in PickyBPE to 0.9 as used in the original paper (Chizhov et al., 2024). Similarly, for OBPE, we use the default hyperparameter values. For transliteration of native Indic script to the ISO-15919 script, we used `https://www.aksharamukha.com/`.

| Non-Normalized | Normalized |
|---|---|
| गांधीनगर<br>(gāṃdhīnagara) | गान्धीनगर<br>(gāndhīnagara) |
| गंगातट<br>(gaṃgātaṭa) | गङ्गातट<br>(gaṅgātaṭa) |
| चंद्रवदन<br>(caṃdravadana) | चन्द्रवदन<br>(candravadana) |
| चंद्रवंशी<br>(caṃdravaṃśī) | चन्द्रवंशी<br>(candravaṃśī) |
| पंकजनयना<br>(paṃkajanayanā) | पङ्कजनयना<br>(paṅkajanayanā) |
| जगदंबा<br>(jagadaṃbā) | जगदम्बा<br>(jagadambā) |
| कुंभमेला<br>(kuṃbhamelā) | कुम्भमेला<br>(kumbhamelā) |

Figure 3: Illustration of fifth-case normalization for Devanagari script. The left column shows non-normalized words containing variant Unicode characters and diacritics, while the right column presents their normalized equivalents after applying custom anusvāra rules. Bracketed text contains transliteration in ISO-15919 format for readability.

## C Experimental Setup

**Pre-training & Finetuning:** The hyperparameters used for pre-training are detailed in Table 11. Table 12 shows the dataset details of the IndicXtreme benchmark.

## D Clusters

The clusters formed for Native and ISO scripts are presented in Table 13 and 14, respectively. $\text{Cluster}_5$, $\text{Cluster}_6$, $\text{Cluster}_7$, $\text{Cluster}_8$ are formed automatically using Algorithm 1. $\text{Cluster}_{lf}$ depicts a language grouping formed by manually grouping languages belonging to the same language family.

## E Results

### E.1 Instrinsic metrics

#### E.1.1 Normalization

The detailed fertility scores for non-normalized and normalized training corpora for 32k, 64k, 128k, and 256k vocabulary are presented in Table 16.

#### E.1.2 Vocabulary size

The overlap between different pairs of tokenizers across vocabulary sizes are illustrated in Table 18. The overlap increases with an increase in vocabu-lary size up to 128k, after which, we see a reduction in the overlap as seen in Section 18.

| Language | Code | LF | Script | # Rows (M) | # Filtered (M) | 10% Sub-sampled (M) | # Training Corpus (M) |
|---|---|---|---|---|---|---|---|
| Hindi | hin | Indo-European | Devanagari | 17.42 | 15.15 | 1.52 | 0.74 |
| Assamese | asm | Indo-European | Assamese | 0.33 | 0.28 | 0.03 | 0.22 |
| Bengali | ben | Indo-European | Bengali | 11.50 | 10.66 | 1.07 | 0.67 |
| Konkani | gom | Indo-European | Devanagari | 0.01 | 0.01 | 0.00 | 0.08 |
| Gujarati | guj | Indo-European | Gujarati | 3.97 | 3.57 | 0.36 | 0.48 |
| Kannada | kan | Dravidian | Kannada | 3.63 | 3.15 | 0.32 | 0.46 |
| Maithili | mai | Indo-European | Devanagari | 0.02 | 0.02 | 0.00 | 0.10 |
| Malayalam | mal | Dravidian | Malayalam | 6.37 | 5.99 | 0.60 | 0.56 |
| Marathi | mar | Indo-European | Devanagari | 5.87 | 4.99 | 0.50 | 0.53 |
| Nepali | nep | Indo-European | Devanagari | 8.59 | 8.37 | 0.84 | 0.62 |
| Oriya | ori | Indo-European | Odia | 2.00 | 1.90 | 0.19 | 0.40 |
| Punjabi | pan | Indo-European | Gurmuki | 1.74 | 1.50 | 0.15 | 0.37 |
| Sanskrit | san | Indo-European | Devanagari | 0.91 | 0.83 | 0.08 | 0.31 |
| Sindhi | snd | Indo-European | Arabic | 0.54 | 0.40 | 0.04 | 0.25 |
| Tamil | tam | Dravidian | Tamil | 7.83 | 6.47 | 0.65 | 0.57 |
| Urdu | urd | Indo-European | Arabic | 5.44 | 5.17 | 0.52 | 0.54 |
| Telugu | tel | Dravidian | Telugu | 7.08 | 6.10 | 0.61 | 0.56 |
| **Total** | | | | | | | **7.46** |

Table 9: Per-language statistics for tokenizer training data: number of raw and filtered rows, 10% sub-sampled entries, and final corpus sizes in millions (M).

| Hyper-parameter | Value(s) |
|---|---|
| model_type | BPE \| Unigram |
| vocab_size | 32k \| 64k \| 128k \| 256k |
| split_by_unicode_script | True |
| split_by_number | True |
| split_by_whitespace | True |
| split_digits | False |
| train_extremely_large_corpus | True |

Table 10: SentencePiece settings used for training tokenizers. All other options or flags are the default values.

### E.1.3 BPE vs. ULM

Table 17 shows the performance of BPE and ULM across vocabulary sizes of 32k, 64k, 128k, and 256k.

## E.2 BPE vs. PickyBPE

Table 19 reports Fertility and CPT scores across 17 languages comparing BPE and PickyBPE tokenizers with a vocabulary size of 128k. We evaluate the fertility on the FLORES-200 dev set (Costa-Jussà et al., 2022).

# F Extrinsic Evaluation

## F.1 BPE vs. ULM

For extrinsic evaluation, we assess the impact of the tokenization strategy on Named Entity Recognition (NER) performance using an LSTM-based model trained with vocabularies of size 128k. The ULM-based model attains a slightly higher average macro F1 score across languages (72.3), compared to the BPE-based model (71.9), with particularly notable gains in Location (LOC) and Person (PER) entities. In contrast, the model trained using BPE yields marginally better performance on Organization (ORG) entities. Figure 4 summarizes these results, highlighting the consistent advantage of ULM across most entity types. We also observe that results are sensitive to vocabulary size.

| Hyperparameter | Value(s) |
|---|---|
| Train / Eval batch size | 32 / 8 |
| Learning rate | $5\times10^{-5}$ |
| Total training steps | 200,000 |
| Warmup steps | 10,000 |
| Checkpoint interval | 1,000 steps |
| Iterations per loop | 1,000 |
| Vocabulary size | 128,000 |
| Max position embeddings | 512 |
| Attention and Hidden dropout | 0.1 |
| Activation function | GELU |
| Initialization range | 0.02 |
| Optimizer | Adam with weight decay |
| Weight decay | 0.01 |
| $\beta_1$ / $\beta_2$ | 0.9 / 0.999 |
| $\epsilon$ | $1\times10^{-6}$ |
| Gradient clipping | 1.0 |

Table 11: BERT-based pre-training configuration and hyperparameters.

Table 20 reports the average F1 scores and bootstrap means across three random seed runs. The observed differences between tokenizers are marginal and unlikely to be statistically significant, a trend that is consistently reflected in the bootstrap esti-mates.

| Task Category | Dataset | Task Description | Languages | Metric |
|---|---|---|---|---|
| Natural Language Understanding | Amazon Massive | Intent and Slot filling | 5 Indic | F1 |
| Sentence Classification | Indic COPA | Multiple-Choice | 19 Indic + En | Acc. |
| Sentence Classification | Indic Sentiment | Sentiment Analysis | 11 Indic + En | Acc. |
| Sentence Classification | XNLI | Text-Classification | 11 Indic | F1 |
| Sentence Classification | Indic XParaphrase | Binary-Classification | 11 Indic | F1 |
| Question Answering | IndicQA | Extractive Question Answering | 11 Indic | F1 |
| Information Retrieval | Flores-200 | Cross-lingual Retrieval | 14 Indic | Acc. |
| Structure Prediction | Naamapadam | Named Entity Recognition | 12 Indic | F1 |

Table 12: Datasets included in the IndicXTREME benchmark.

| Cluster | Language grouping | Size |
|---|---|---|
| $\text{Cluster}_5$ | (mar), (san, urd, snd, ben, guj, tam, kan, asm, pan, tel, hin, ori), (mai), (nep), (gom) | 5 |
| $\text{Cluster}_6$ | (mar), (ben, asm), (urd, mai, snd, guj, tam, kan, pan, tel, hin, ori), (nep), (gom), (san) | 6 |
| $\text{Cluster}_7$ | (mar), (guj, tam, kan, asm, pan, tel, hin, ori), (mai), (nep), (gom), (san), (urd, snd, ben) | 7 |
| $\text{Cluster}_8$ | (mar), (tam, kan, asm, pan, tel, hin, ori), (mai), (nep), (urd, snd, ben), (gom), (san), (guj) | 8 |
| $\text{Cluster}_{lf}$ | (ben, ori, asm, hin, nep, urd, pan, san, guj, snd, mai, gom, mar), (kan, tel, mal, tam) | 2 |

Table 13: Clusters formed using monolingual corpus in Native for all considered languages

| Cluster | Language grouping | Size |
|---|---|---|
| $\text{Cluster}_5$ | (ben, ori, asm), (kan, snd, hin, nep, urd, pan, tel, san), (mar), (gom, mal, mai, tam), (guj) | 5 |
| $\text{Cluster}_6$ | (ben, asm), (kan, snd, hin, nep, urd, pan, tel, san), (mar), (gom, mal, mai, tam), (guj), (ori) | 6 |
| $\text{Cluster}_7$ | (ben, asm), (kan, snd, hin, urd, pan, tel, san), (mar), (gom, mal, mai, tam), (guj), (ori), (nep) | 7 |
| $\text{Cluster}_8$ | (ben), (kan, snd, hin, urd, pan, tel, san), (mar), (gom, mal, mai, tam), (guj), (ori), (nep), (asm) | 8 |
| $\text{Cluster}_{lf}$ | (ben, ori, asm, hin, nep, urd, pan, san, guj, snd, mai, gom, mar), (kan, tel, mal, tam) | 2 |

Table 14: Clusters formed using monolingual corpus in ISO 15919 for all considered languages

| Language | # words | Normalized word count (%) | # unique words | Normalized Unique word count (%) |
|---|---|---|---|---|
| Hindi | 24607 | 1123 (4.56) | 6600 | 604 (9.15) |
| Assamese | 18570 | 154 (0.83) | 8320 | 118 (1.42) |
| Bengali | 18756 | 103 (0.55) | 7801 | 72 (0.92) |
| Konkani | 25322 | 1417 (5.6) | 6979 | 638 (9.14) |
| Gujarati | 20080 | 1033 (5.14) | 8701 | 697 (8.01) |
| Kannada | 15430 | 773 (5.01) | 9730 | 628 (6.45) |
| Maithili | 23455 | 1196 (5.1) | 6832 | 621 (9.09) |
| Malayalam | 14377 | 283 (1.97) | 9494 | 242 (2.55) |
| Marathi | 18065 | 1937 (10.72) | 9560 | 1264 (13.22) |
| Nepali | 17847 | 93 (0.52) | 7811 | 60 (0.77) |
| Oriya | 18914 | 71 (0.38) | 7480 | 40 (0.53) |
| Punjabi | 24539 | 611 (2.49) | 6668 | 260 (3.90) |
| Sanskrit | 16671 | 480 (2.88) | 9187 | 390 (4.25) |
| Sindhi | 23345 | 876 (3.75) | 9534 | 481 (5.05) |
| Tamil | 16134 | 0 (0.0) | 9147 | 0 (0.00) |
| Telugu | 16388 | 3471 (21.18) | 10421 | 1968 (18.88) |
| **Total** | **312500** | **13621 (4.36)** | **134265** | **8083 (6.02)** |

Table 15: Effect of fifth-case normalization on Flores *devtest* data

### F.2 Cluster vs. ULM

We have conducted extensive experiments for the evaluation of the $\text{Cluster}_{lf}$ and ULM tokenisers on the IndicXTREME[6] benchmark. We evaluated using a BERT-Base architecture with 12 transformer layers, hidden size 768, and 768-dimensional sentence embeddings. All experiments are conducted on 8 A6000 GPU. To ensure robustness, zero-shot evaluation is performed across five random seeds 42, 45, 18, 10, and 7 for all datasets except Amazon Massive. For the Amazon Massive dataset, we have conducted zero-shot evaluations across random seeds 18, 42, 45.

[6] Link for IndicXTREME benchmark: https://huggingface.co/collections/ai4bharat/indicxtreme

| Lang. | Algorithm | Vocab | | | | | | | |
|---|---|---|---|---|---|---|---|---|---|
| | | 32k | | 64k | | 128k | | 256k | |
| | | NN | N | NN | N | NN | N | NN | N |
| hin | BPE | 1.533 | 1.523 | 1.404 | 1.309 | 1.309 | 1.301 | 1.244 | 1.236 |
| | ULM | 1.517 | 1.509 | 1.394 | 1.387 | 1.303 | 1.296 | 1.257 | 1.252 |
| mai | BPE | 1.655 | 1.649 | 1.484 | 1.477 | 1.378 | 1.373 | 1.307 | 1.302 |
| | ULM | 1.681 | 1.674 | 1.524 | 1.518 | 1.401 | 1.399 | 1.320 | 1.317 |
| mar | BPE | 2.107 | 1.998 | 1.785 | 1.770 | 1.698 | 1.593 | 1.473 | 1.462 |
| | ULM | 1.963 | 1.956 | 1.746 | 1.733 | 1.573 | 1.566 | 1.482 | 1.472 |
| npi | BPE | 1.955 | 1.924 | 1.735 | 1.710 | 1.576 | 1.557 | 1.450 | 1.435 |
| | ULM | 1.921 | 1.895 | 1.717 | 1.697 | 1.565 | 1.549 | 1.470 | 1.457 |
| gom | BPE | 2.376 | 2.378 | 2.082 | 2.086 | 1.854 | 1.853 | 1.673 | 1.671 |
| | ULM | 2.337 | 2.342 | 2.068 | 2.064 | 1.815 | 1.813 | 1.685 | 1.684 |
| san | BPE | 2.446 | 2.428 | 2.206 | 2.180 | 2.011 | 1.994 | 1.862 | 1.845 |
| | ULM | 2.418 | 2.400 | 2.186 | 2.170 | 1.986 | 1.971 | 1.840 | 1.831 |
| snd | BPE | 2.327 | 2.347 | 2.182 | 2.191 | 2.029 | 2.028 | 1.905 | 1.908 |
| | ULM | 2.327 | 2.351 | 2.184 | 2.203 | 1.992 | 2.024 | 1.875 | 1.892 |
| pan | BPE | 1.829 | 1.825 | 1.595 | 1.590 | 1.435 | 1.433 | 1.331 | 1.329 |
| | ULM | 1.776 | 1.774 | 1.561 | 1.558 | 1.420 | 1.417 | 1.363 | 1.363 |
| ben | BPE | 1.937 | 1.935 | 1.692 | 1.689 | 1.506 | 1.503 | 1.390 | 1.387 |
| | ULM | 1.872 | 1.878 | 1.659 | 1.657 | 1.489 | 1.487 | 1.393 | 1.393 |
| asm | BPE | 2.285 | 2.278 | 1.992 | 1.988 | 1.757 | 1.752 | 1.620 | 1.598 |
| | ULM | 2.244 | 2.235 | 1.956 | 1.951 | 1.744 | 1.740 | 1.618 | 1.617 |
| kan | BPE | 2.761 | 2.745 | 2.367 | 2.358 | 2.048 | 2.034 | 1.800 | 1.788 |
| | ULM | 2.699 | 2.680 | 2.309 | 2.294 | 1.993 | 1.997 | 1.801 | 1.786 |
| tel | BPE | 2.580 | 2.571 | 2.205 | 2.201 | 1.897 | 1.889 | 1.681 | 1.676 |
| | ULM | 2.546 | 2.531 | 2.164 | 2.152 | 1.870 | 1.863 | 1.701 | 1.693 |
| mal | BPE | 3.075 | 3.052 | 2.645 | 2.616 | 2.280 | 2.241 | 1.995 | 1.957 |
| | ULM | 3.014 | 2.983 | 2.581 | 2.552 | 2.244 | 2.202 | 1.993 | 1.954 |
| tam | BPE | 2.488 | 2.485 | 2.158 | 2.156 | 1.884 | 1.882 | 1.691 | 1.690 |
| | ULM | 2.400 | 2.399 | 2.075 | 2.073 | 1.840 | 1.836 | 1.675 | 1.677 |
| guj | BPE | 2.067 | 2.067 | 1.798 | 1.797 | 1.599 | 1.595 | 1.457 | 1.455 |
| | ULM | 2.000 | 1.995 | 1.755 | 1.752 | 1.840 | 1.836 | 1.675 | 1.677 |
| ori | BPE | 2.284 | 2.264 | 1.960 | 1.942 | 1.703 | 1.683 | 1.534 | 1.515 |
| | ULM | 2.240 | 2.217 | 1.912 | 1.891 | 1.680 | 1.655 | 1.550 | 1.532 |
| urd | BPE | 1.619 | 1.474 | 1.453 | 1.321 | 1.330 | 1.207 | 1.251 | 1.136 |
| | ULM | 1.568 | 1.432 | 1.424 | 1.295 | 1.316 | 1.197 | 1.278 | 1.164 |

Table 16: Fertility scores comparison between normalized and non-normalized text. Here, NN and N represent Non-normalized and normalized, respectively.

| Algorithm | Vocab | | | | | | | | | | | |
|---|---|---|---|---|---|---|---|---|---|---|---|---|
| | 32k | | | 64k | | | 128k | | | 256k | | |
| | F (↓) | CPT (↑) | WFR (↓) | F (↓) | CPT (↑) | WFR (↓) | F (↓) | CPT (↑) | WFR (↓) | F (↓) | CPT (↑) | WFR (↓) |
| BPE | 2.137 | 3.178 | 57.280 | 1.872 | 3.619 | 48.496 | 1.664 | 4.059 | 40.509 | 1.517 | 4.444 | 33.879 |
| ULM | 2.093 | 3.243 | 54.679 | 1.838 | 3.682 | 46.844 | 1.642 | 4.110 | 40.049 | 1.530 | 4.405 | 37.141 |

Table 17: Average fertility (F), character per token (CPT), and word fragmentation rate (WFR) across in a joint setting. Lower fertility and WFR indicate better segmentation quality. Higher CPT suggests tokens are more semantically meaningful and compact.

| Algorithm | Vocabulary Overlap (%) | | | |
|---|---|---|---|---|
| | 32k | 64k | 128k | 256k |
| ULM (NN) vs BPE (NN) | 65.40 | 68.19 | 74.32 | 68.35 |
| ULM (N) vs BPE (N) | 65.48 | 68.15 | 74.28 | 67.74 |
| BPE (N) vs BPE (NN) | 92.95 | 92.72 | 92.25 | 91.95 |
| ULM (N) vs ULM (NN) | 93.35 | 93.14 | 92.87 | 92.36 |

Table 18: Percentage of vocabulary overlap across tokenizers. Here, NN and N represent non-normalized and normalized text, respectively.

| Lang. | Fertility (↓) BPE | PBPE | $\Delta_F$ | CPT (↑) BPE | PBPE | $\Delta_{CPT}$ |
|---|---|---|---|---|---|---|
| asm | 1.752 | 1.675 | 0.077 | 3.801 | 3.979 | 0.178 |
| ben | 1.503 | 1.451 | 0.052 | 4.455 | 4.617 | 0.162 |
| gom | 1.853 | 1.805 | 0.048 | 3.808 | 3.913 | 0.105 |
| guj | 1.595 | 1.528 | 0.067 | 3.817 | 3.986 | 0.169 |
| hin | 1.301 | 1.232 | 0.068 | 3.987 | 4.209 | 0.222 |
| kan | 2.034 | 1.984 | 0.051 | 4.271 | 4.380 | 0.109 |
| mai | 1.373 | 1.315 | 0.058 | 3.884 | 4.057 | 0.173 |
| mal | 2.241 | 2.186 | 0.055 | 4.431 | 4.542 | 0.111 |
| mar | 1.593 | 1.531 | 0.062 | 4.469 | 4.655 | 0.186 |
| nep | 1.557 | 1.495 | 0.061 | 4.386 | 4.566 | 0.181 |
| ori | 1.683 | 1.650 | 0.034 | 4.086 | 4.172 | 0.085 |
| pan | 1.433 | 1.356 | 0.077 | 3.671 | 3.833 | 0.211 |
| san | 1.994 | 1.969 | 0.025 | 3.762 | 3.813 | 0.051 |
| snd | 1.393 | 1.335 | 0.062 | 3.422 | 3.587 | 0.164 |
| tam | 1.882 | 1.814 | 0.068 | 4.841 | 5.022 | 0.181 |
| tel | 1.889 | 1.842 | 0.047 | 4.242 | 4.376 | 0.135 |
| urd | 1.207 | 1.179 | 0.028 | 3.665 | 3.751 | 0.086 |
| **Avg.** | **1.664** | **1.609** | **0.055** | **4.059** | **4.209** | **0.148** |

Table 19: Fertility and CPT scores across 17 languages compared for BPE and PBPE with vocabulary size of 128k. $\Delta_F$ = Fertility BPE − Fertility PBPE. $\Delta_{CPT}$ = CPT PBPE - CPT BPE.

| Algorithm | Average Macro F1 | Bootstrap Mean |
|---|---|---|
| **BPE** | 0.7229 | 0.7270 |
| **ULM** | 0.7235 | 0.7273 |

Table 20: Average Macro F1 and Bootstrap mean run across three different seeds for the NER task.

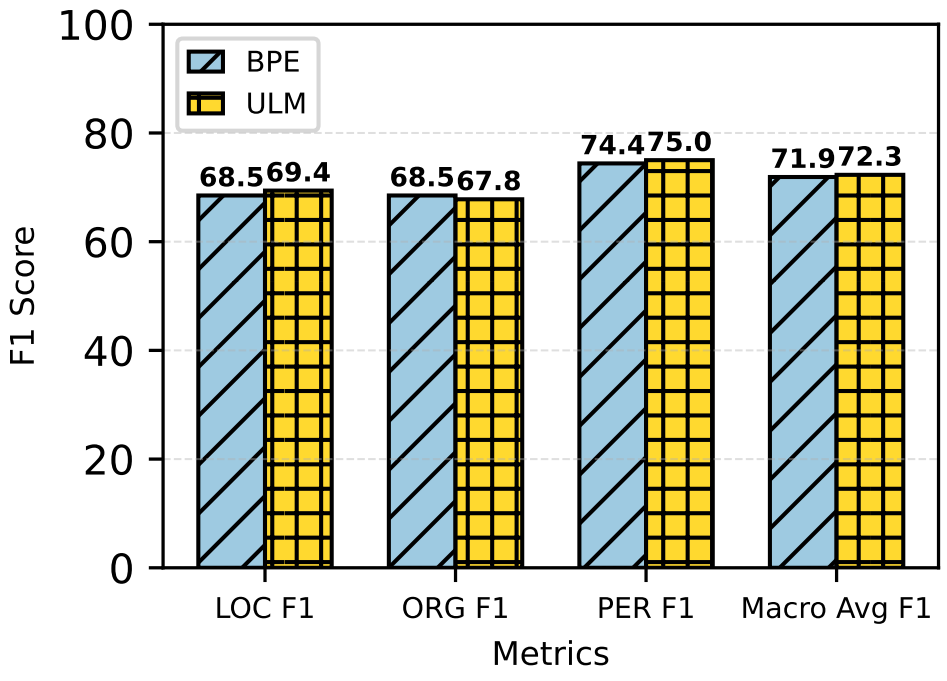

Figure 4: Comparison for BPE and ULM with vocabulary size of 128k trained to perform the NER task. Trained on a language-specific script, normalized data.

### F.2.1 Retrieval on FLORES Dataset

For cross-lingual evaluation, we use sentence-aligned data from the FLORES-200 devtest split, covering 13 Indic-to-Hindi language pairs. For each source–target pair, embeddings are independently computed for both languages. Cross-lingual retrieval is performed using FAISS IndexFlatL2, with optional L2-normalization when cosine similarity is selected. In the performed zero-shot experimentation, similarity is computed using cosine distance, implemented via vector normalization followed by nearest-neighbour search. **Retrieval accuracy** is measured as top-1 alignment accuracy, where a prediction is considered correct if the nearest target embedding corresponds to the aligned sentence index. Overall, Table 21 reports ULM performs better than $\text{Cluster}_{lf}$.

| Src. Lang. | $\text{Cluster}_{lf}$ | ULM |
|---|---|---|
| asm | 2.27 ± 0.00 | **3.06 ± 0.00** |
| ben | **7.31 ± 0.00** | 5.43 ± 0.00 |
| guj | 1.38 ± 0.00 | **1.58 ± 0.00** |
| kan | 2.27 ± 0.00 | **2.47 ± 0.00** |
| mai | 28.95 ± 0.00 | **31.82 ± 0.00** |
| mal | 2.17 ± 0.00 | **3.26 ± 0.00** |
| mar | **28.56 ± 0.00** | 27.37 ± 0.00 |
| npi | 30.24 ± 0.00 | **33.70 ± 0.00** |
| ory | **1.28 ± 0.00** | 0.89 ± 0.00 |
| pan | **11.36 ± 0.00** | 9.88 ± 0.00 |
| san | **10.08 ± 0.00** | 8.99 ± 0.00 |
| tam | **6.42 ± 0.00** | 5.73 ± 0.00 |
| tel | **3.75 ± 0.00** | **3.75 ± 0.00** |
| Avg. | 16.86 ± 0.00 | **16.99 ± 0.00** |

Table 21: Zero-shot retrieval accuracy scores on FLORES dataset taking Hindi as a reference language.

Input sentences are tokenized with a maximum sequence length of 128, padding to max length, and truncation enabled. Embeddings are computed in batches of 32 sentences, and sentence-level representations are obtained using mean pooling over token embeddings.

### F.2.2 IndicQA Dataset

For the Question-Answering experiments, maximum input sequence length of 512 tokens and a maximum question length of 64 tokens is set. A document stride of 128 tokens is applied to handle long contexts. Answer spans are selected based on the maximum scoring span, with confidence computed as the softmax probability of the chosen span and batch size of 8 is defined. Table 22 shows the $\text{Cluster}_{lf}$ performs better than ULM.

### F.2.3 IndicCOPA Dataset

Table 23 shows ULM performs better than $\text{Cluster}_{lf}$ under zero-shot settings. For each input, the premise is paired with two hypothesis templates conditioned on the question type (cause or effect), and the prediction is selected based on the highest entailment score returned by the zero-shot classifier. Batch size is kept as 1.

### F.2.4 IndicXParaphrase Dataset

All experiments are conducted with a batch size and maximum input sequence length of 512 to-kens, where truncation is applied and padding is performed at the sentence level.
For classification, cosine similarity scores are thresholded to obtain final predictions. A default threshold of 0.5 is used for primary evaluation. Additionally, we perform threshold tuning for each language by sweeping values from 0.1 to 0.95 in increments of 0.05, selecting the threshold that maximizes the weighted F1-score on the validation set. Overall, Table 24 shows that ULM performs better than Cluster$_{lf}$.

| Lang. | **Cluster$_{lf}$** | **ULM** |
|---|---|---|
| tam | **2.25** ± **0.52** | 1.68 ± 0.30 |
| kan | 2.71 ± 0.59 | **3.40** ± **0.55** |
| hin | **2.08** ± **0.59** | 1.63 ± 0.30 |
| mal | 1.63 ± 0.36 | **1.77** ± **0.30** |
| guj | 2.88 ± 1.00 | **3.07** ± **0.38** |
| ben | **2.57** ± **0.46** | 2.17 ± 0.31 |
| ori | **3.50** ± **0.81** | 3.49 ± 0.34 |
| asm | **3.05** ± **0.24** | 2.76 ± 0.57 |
| mar | **3.27** ± **0.70** | 2.68 ± 0.46 |
| tel | 1.50 ± 0.44 | **1.51** ± **0.25** |
| pan | **3.19** ± **0.47** | **3.19** ± **0.08** |
| Avg. | **2.60** ± **0.56** | 2.49 ± 0.35 |

Table 22: F1 scores of the zero-shot performance on IndicQA dataset.

| Lang. | **Cluster$_{lf}$** | **ULM** |
|---|---|---|
| tam | **51.40** ± **1.96** | 48.36 ± 2.79 |
| sat | 51.36 ± 2.43 | **52.24** ± **1.28** |
| mai | 50.48 ± 2.09 | **51.16** ± **1.08** |
| gom | 50.24 ± 3.24 | **50.32** ± **1.43** |
| npi | 49.76 ± 2.33 | **50.68** ± **4.22** |
| eng | **49.72** ± **1.75** | 49.52 ± 3.71 |
| mal | 49.48 ± 3.42 | **49.76** ± **2.24** |
| kan | **49.40** ± **2.53** | 49.08 ± 1.59 |
| tel | 49.32 ± 2.75 | **50.28** ± **3.10** |
| ben | 49.16 ± 1.35 | **50.92** ± **1.00** |
| pan | 49.16 ± 2.50 | **50.16** ± **2.67** |
| san | 49.12 ± 1.82 | **50.80** ± **1.61** |
| ori | **49.12** ± **3.74** | 48.56 ± 2.08 |
| asm | 48.88 ± 1.63 | **49.36** ± **2.06** |
| mar | 48.73 ± 2.40 | **50.78** ± **1.88** |
| hin | 48.55 ± 1.51 | **50.38** ± **1.94** |
| guj | 48.30 ± 2.67 | **50.89** ± **4.17** |
| Avg. | 49.54 ± 2.36 | **50.19** ± **2.29** |

Table 23: Accuracy scores of the zero-shot performance on IndicCOPA dataset.

#### F.2.5 IndicSentiment Dataset

An evaluation is performed with a batch size of 16 and without data shuffling, using a custom collation function. Input texts are tokenized with both padding and truncation enabled, capped at a maximum sequence length of 512 tokens.

| Lang. | **Cluster$_{lf}$** | **ULM** |
|---|---|---|
| asm | 33.95 ± 0.00 | **34.27** ± **0.00** |
| ben | **41.45** ± **0.00** | 41.37 ± 0.00 |
| guj | **49.45** ± **0.00** | 48.78 ± 0.00 |
| hin | 36.90 ± 0.00 | **37.36** ± **0.00** |
| kan | 33.27 ± 0.00 | **33.33** ± **0.00** |
| mal | 33.00 ± 0.00 | **33.38** ± **0.00** |
| mar | **33.81** ± **0.00** | 33.80 ± 0.00 |
| ori | **33.33** ± **0.00** | **33.33** ± **0.00** |
| pan | 71.37 ± 0.00 | **71.42** ± **0.00** |
| tel | 43.97 ± 0.00 | **44.52** ± **0.00** |
| Avg. | 41.05 ± 0.00 | **41.16** ± **0.00** |

Table 24: F1 scores of the zero-shot performance on IndicXParaphrase dataset.

| Lang. | **Cluster$_{lf}$** | **ULM** |
|---|---|---|
| asm | 48.88 ± 1.63 | **49.36** ± **2.06** |
| ben | 49.16 ± 1.35 | **50.92** ± **1.00** |
| eng | **49.72** ± **1.75** | 49.52 ± 3.71 |
| gom | 50.24 ± 3.24 | **50.32** ± **1.43** |
| guj | 48.30 ± 2.67 | **50.89** ± **4.17** |
| hin | 48.55 ± 1.51 | **50.38** ± **1.94** |
| kan | **49.40** ± **2.53** | 49.08 ± 1.59 |
| mai | 50.48 ± 2.09 | **51.16** ± **1.08** |
| mal | 49.48 ± 3.42 | **49.76** ± **2.24** |
| mar | 48.73 ± 2.40 | **50.78** ± **1.88** |
| npi | 49.76 ± 2.33 | **50.68** ± **4.22** |
| ori | **49.12** ± **3.74** | 48.56 ± 2.08 |
| pan | 49.16 ± 2.50 | **50.16** ± **2.67** |
| san | 49.12 ± 1.82 | **50.80** ± **1.61** |
| sat | 51.36 ± 2.43 | **52.24** ± **1.28** |
| tam | **51.40** ± **1.96** | 48.36 ± 2.79 |
| tel | 49.32 ± 2.75 | **50.28** ± **3.10** |
| Avg. | 49.54 ± 2.36 | **50.19** ± **2.29** |

Table 25: Accuracy scores on the zero-shot performance on IndicSentiment dataset.

Overall, Table 25 shows that ULM performs better than Cluster$_{lf}$ under zero-shot settings.

#### F.2.6 Amazon Massive Dataset

**Zero-shot Evaluation Details:** Evaluations are performed with a maximum sequence length of 128 and greedy argmax decoding. Table 26 reports macro-F1 scores for intent detection and slot-filling. Cluster$_{lf}$ performs better for intent detection and ULM for the slot-filling task.

**Supervised Fine-tuning Details:** The joint intent classification and slot-filling model is fine-tuned using a shared multilingual BERT encoder with two task-specific linear heads. Input sequences are tokenized to a maximum length of 128 tokens, with padding or truncation applied as required. Training is performed using the AdamW optimizer with a learning rate of $2 \times 10^{-5}$, weight decay of 0.01, and 500 warmup steps. A per-device batch size of 16 is used for both training and evaluation. The model is trained for up to 100 epochs with early stopping (patience = 3) based on validation joint F1, defined as the average of intent F1 and slot F1. For intent classification, cross-entropy loss is computed over pooled [CLS] representations, while slot filling uses token-level cross-entropy loss with ignored indices ($-100$) for special tokens and subword continuations. The final objective is the unweighted sum of intent and slot losses. All experiments are conducted with a random seed of 42. Table 27 reports Cluster$_{lf}$ performs better for intent detection and ULM for the slot-filling task.

| | Intent | | Slot | |
|---|---|---|---|---|
| **Lang.** | **Cluster$_{lf}$** | **ULM** | **Cluster$_{lf}$** | **ULM** |
| hin | **0.19 ± 0.07** | 0.11 ± 0.07 | 0.00 ± 0.00 | **0.14 ± 0.11** |
| kan | **0.13 ± 0.05** | **0.13 ± 0.07** | 0.00 ± 0.00 | **0.21 ± 0.29** |
| mal | **0.23 ± 0.17** | 0.20 ± 0.08 | 0.00 ± 0.00 | **0.57 ± 0.81** |
| tam | 0.12 ± 0.02 | **0.14 ± 0.09** | 0.00 ± 0.00 | **0.07 ± 0.10** |
| tel | 0.12 ± 0.08 | **0.13 ± 0.09** | 0.00 ± 0.00 | **0.04 ± 0.06** |
| AVG | **0.16 ± 0.08** | 0.14 ± 0.08 | 0.00 ± 0.00 | **0.21 ± 0.27** |

Table 26: Macro F1 scores on the zero-shot intent detection and slot filling performance on Amazon Massive dataset.

| | Intent | | Slot | |
|---|---|---|---|---|
| **Lang.** | **Cluster$_{lf}$** | **ULM** | **Cluster$_{lf}$** | **ULM** |
| hin | **72.98 ± 0.36** | 72.43 ± 0.21 | **8.50 ± 8.33** | 0.17 ± 0.00 |
| kan | **72.52 ± 0.34** | 72.23 ± 0.72 | **5.76 ± 5.07** | 0.23 ± 0.00 |
| mal | 73.02 ± 0.75 | **73.52 ± 0.61** | 3.94 ± 6.40 | **11.36 ± 10.51** |
| tam | **72.98 ± 0.57** | 72.40 ± 0.59 | 2.76 ± 2.40 | **5.29 ± 0.00** |
| tel | 72.20 ± 1.04 | **72.54 ± 1.11** | 0.22 ± 0.05 | **8.18 ± 7.30** |
| Avg. | **72.74 ± 0.61** | 72.62 ± 0.65 | 4.24 ± 4.45 | **5.04 ± 3.56** |

Table 27: Macro F1 scores on the fine-tuned intent classification and slot filling performance on the Amazon Massive dataset.

#### F.2.7 IndicXNLI Dataset

**Zero-shot Evaluation:** For zero-shot evaluation, a batch size of 32 and maximum input length of 128 is used. Table 28 reports that ULM consistenly performs better than Cluster$_{lf}$.

**Supervised Fine-tuning Details:** Fine-tuning of the BERT-base model is performed for three-way classification: entailment, neutral, and contradiction. Input sentence pairs (premise, hypothesis) are tokenized with a maximum sequence length of 128, applying padding to the maximum length and truncation. Training is performed with a per-device batch size of 32, learning rate of $2 \times 10^{-5}$, weight decay of 0.01, and AdamW optimization. We train for up to 100 epochs with linear learning-rate warmup over 500 steps, enabling FP16 mixed-precision training on GPUs. Early stopping is applied with a patience of 3 evaluation steps with random seed of 42. Table 29 reports that Cluster$_{lf}$ consistently outperforms the ULM after the optimal fine-tuning is done.

| **Lang.** | **Cluster$_{lf}$** | **ULM** |
|---|---|---|
| asm | 21.30 ± 0.00 | **26.18 ± 0.00** |
| ben | 22.07 ± 0.00 | **26.44 ± 0.00** |
| guj | 22.78 ± 0.00 | **26.85 ± 0.00** |
| hin | 24.25 ± 0.00 | **26.90 ± 0.00** |
| kan | 24.09 ± 0.00 | **26.87 ± 0.00** |
| mal | 25.28 ± 0.00 | **26.97 ± 0.00** |
| mar | 24.14 ± 0.00 | **26.69 ± 0.00** |
| ori | 21.93 ± 0.00 | **24.97 ± 0.00** |
| pan | 21.59 ± 0.00 | **26.78 ± 0.00** |
| tam | 22.36 ± 0.00 | **26.70 ± 0.00** |
| tel | 22.77 ± 0.00 | **26.90 ± 0.00** |
| Avg. | 22.96 ± 0.00 | **26.66 ± 0.00** |

Table 28: F1 scores on the zero-shot performance on IndicXNLI dataset.

| **Lang.** | **Cluster$_{lf}$** | **ULM** |
|---|---|---|
| asm | **64.93** | 63.98 |
| ben | 65.91 | **66.14** |
| guj | **65.12** | 64.12 |
| hin | **67.92** | 67.05 |
| kan | **65.80** | 64.08 |
| mal | **64.69** | 62.29 |
| mar | **65.00** | 63.52 |
| ori | **65.36** | 64.05 |
| pan | **67.15** | 65.44 |
| tam | **65.22** | 63.61 |
| tel | **64.99** | 63.58 |
| Avg. | **65.65** | 64.42 |

Table 29: F1 scores on the fine-tuned XNLI performance across Indic languages.

#### F.2.8 NaamaPadam Structure Prediction

**Zero-shot Evaluation:** For zero-shot evaluation, a per-device batch size of 32, maximum sequence length of 128 tokens, and enable mixed-precision inference. Table 30 reports that overall ULM performs better than Cluster$_{lf}$.

**Supervised Fine-tuning Details:** Fine-tuning is performed on the normalized native-script Naamapadam dataset spanning 11 Indic languages. Input sequences are tokenized with a maximum length of 128 tokens. A per-device batch size of 16 for both training and evaluation, a learning rate of $5 \times 10^{-5}$, weight decay of 0.01, and a total of 100 epochs. A linear learning rate scheduler with 50 warmup steps is employed. We enable mixed-precision training for efficiency. Early stopping with a patience of 3 epochs is applied, and the best model is selected based on validation F1-score and a random seed of 42. Table 31 reports that $\text{Cluster}_{lf}$ consistently outperforms the ULM after the optimal fine-tuning is done.

| Lang. | $\textbf{Cluster}_{lf}$ | **ULM** |
|---|---|---|
| asm | 2.03 ± 0.00 | **3.01 ± 0.00** |
| ben | **6.53 ± 0.00** | 5.96 ± 0.00 |
| guj | 4.79 ± 0.00 | **7.18 ± 0.00** |
| hin | 4.38 ± 0.00 | **5.17 ± 0.00** |
| kan | **7.45 ± 0.00** | 5.41 ± 0.00 |
| mal | 6.30 ± 0.00 | **7.10 ± 0.00** |
| mar | **6.95 ± 0.00** | 6.81 ± 0.00 |
| ori | **3.62 ± 0.00** | 2.50 ± 0.00 |
| pan | 4.97 ± 0.00 | **5.97 ± 0.00** |
| tam | **6.59 ± 0.00** | 5.72 ± 0.00 |
| tel | 6.15 ± 0.00 | **6.32 ± 0.00** |
| Avg. | 5.49 ± 0.00 | **5.65 ± 0.00** |

Table 30: F1 scores on the zero-shot structure prediction on NaamaPadam dataset.

| Lang. | $\textbf{Cluster}_{lf}$ | **ULM** |
|---|---|---|
| asm | **56.25** | 48.00 |
| ben | 81.47 | **82.06** |
| guj | 81.24 | **82.77** |
| hin | **83.56** | 82.60 |
| kan | **84.65** | 84.51 |
| mal | 83.58 | **84.20** |
| mar | **83.73** | 82.68 |
| ori | **37.79** | 37.64 |
| pan | **75.92** | 74.83 |
| tam | **76.29** | 73.27 |
| tel | 84.40 | **84.44** |
| Avg. | **75.35** | 74.27 |

Table 31: F1 scores on the fine-tuned structure prediction performance on NaamaPadam dataset.